\documentclass[letterpaper, 10 pt, conference]{ieeeconf}  

\IEEEoverridecommandlockouts                              

\usepackage[utf8]{inputenc}
\usepackage{hyperref}
\usepackage{amsmath}
\usepackage{graphicx}
\usepackage{caption}
\usepackage{subcaption}
\usepackage{multirow}
\usepackage{bm}
\usepackage{rotating}
\usepackage{multirow}
\usepackage{hhline}
\usepackage{verbatim}
\usepackage{soul}
\usepackage{afterpage}
\usepackage{color}
\usepackage{latexsym}
\usepackage{dingbat}
\usepackage{pifont}
\usepackage{amssymb}
\usepackage{amsmath}

\usepackage{pdflscape}
\usepackage{enumerate}
\let\labelindent\relax
\usepackage[inline]{enumitem}
\usepackage{xspace} 
\usepackage{amsmath}
\usepackage[ruled]{algorithm2e}
\usepackage{algorithmicx}
\usepackage{float}
\usepackage[table]{xcolor}
\usepackage[normalem]{ulem}
\usepackage{makecell}

\usepackage{pifont}

\newcolumntype{P}[1]{>{\centering\arraybackslash}p{#1}}
\newcolumntype{M}[1]{>{\centering\arraybackslash}m{#1}}

\SetKwData{Left}{left}\SetKwData{This}{this}\SetKwData{Up}{up}
\SetKwFunction{Union}{Union}\SetKwFunction{FindCompress}{FindCompress}
\SetKwInOut{Input}{input}\SetKwInOut{Output}{output}

\newlength\kwinlength

\newlength\kwoutlength

\usepackage[per=slash]{siunitx}
\usepackage{comment}

\definecolor{OliveGreen}{RGB}{0,200,25}
\newcommand{\red}[1]{\textcolor{red}{#1}}

\newcommand{\darkgreen}[1]{\textcolor{OliveGreen}{#1}}

\newcommand{\ie}{i.\,e.\ }

\newcommand{\armarVI}{\mbox{ARMAR-6}\xspace}

\newcommand{\ackoml}{The research leading to these results has received funding from the German Federal Ministry of Education and Research (BMBF) under the project OML (01IS18040A)}

\newcommand{\replaced}[2]{\red{\ifmmode\text{\sout{\ensuremath{#1}}}\else\sout{#1}\fi}\darkgreen{#2}}
\newcommand{\removed}[1]{\red{\ifmmode\text{\sout{\ensuremath{#1}}}\else\sout{#1}\fi}}

\newcommand{\removedfootnote}[1]{\footnote{\removed{#1}}}
\newcommand{\removedsubsection}[1]{\subsection{\texorpdfstring{\removed{#1}}{#1}}}

\usepackage[per=slash]{siunitx}
\usepackage{amsbsy}

\newcommand{\mq}{\pmb{q}}
\newcommand{\my}{\pmb{y}}

\newcommand{\mz}{\pmb{z}}
\newcommand{\mb}{\pmb{b}}

\newcommand{\mA}{\pmb{A}}

\newcommand{\mSig}{\pmb{\Sigma}}

\title{Learning to Shift Attention for Motion Generation}
\author{You~Zhou, Jianfeng~Gao and~Tamim~Asfour
\thanks{\ackoml}
\thanks{The authors are with the Institute for Anthropomatics and Robotics, Karlsruhe Institute of Technology, Karlsruhe, Germany. {\tt \{you.zhou, asfour\}@kit.edu}}
}

\begin{document}
\inputencoding{utf8}

\maketitle
\thispagestyle{empty}
\pagestyle{empty}


\begin{abstract}
One challenge of motion generation using robot learning from demonstration techniques is that human demonstrations follow a distribution with multiple modes for one task query. Previous approaches fail to capture all modes or tend to average modes of the demonstrations and thus generate invalid trajectories. The other difficulty is the small number of demonstrations that cannot cover the entire working space. To overcome this problem, a motion generation model with extrapolation ability is needed.  Previous works restrict task queries as local frames and learn representations in local frames. We propose a model to solve both problems. For multiple modes, we suggest to learn local latent representations of motion trajectories with a density estimation method based on real-valued non-volume preserving (RealNVP) transformations that provides a set of powerful, stably invertible, and learnable transformations. To improve the extrapolation ability, we propose to shift the attention of the robot from one local frame to another during the task execution.  In experiments, we consider the docking problem used also in previous works where a trajectory has to be generated to connect two dockers without collision. We increase complexity of the task and show that the proposed method outperforms other approaches. In addition, we evaluate the approach in real robot experiments. 
\end{abstract}


\section{Introduction} \label{sec:introduction}

Learning from demonstrations (LfD) is a promising approach in robotics research \cite{Billard2016}. It simplifies robot programming and enables a flexible realization of a variety of robot applications. Two critical challenges of LfD are how to represent motions and how to generalize the learned skill to different task queries.  A proper motion representation simplifies skill generalization and allows generating motion trajectories for inexperienced task queries. To address the problem of skill generalization, we can learn a mapping from task queries to an appropriate motion trajectory based on multiple human demonstrations.
Learning such a mapping is not a trivial task. One challenge is that the demonstrations usually contain multiple modes, which should be considered for learning a motion generation model. However, multiple modes cannot be learned by many regression models such as a Gaussian process regression or a fully-connected neural network.

 Another difficulty is that usually a rather small number of demonstrations is available which do not cover the entire work space. Thus, the learned motion generation model should be able to extrapolate to task queries that are not in the training queries' range.

To meet these two challenges, based on the idea of local frames, we propose a model that consists of 1) a latent transformation that transforms demonstration trajectories to a latent space with a single mode distribution, and 2) a recurrent attention model that shifts the attention of the robot from one local frame to another.

\section{Related Work} \label{sec:related}

\begin{figure}[t]
  \centering
    \includegraphics[width=0.9\columnwidth]{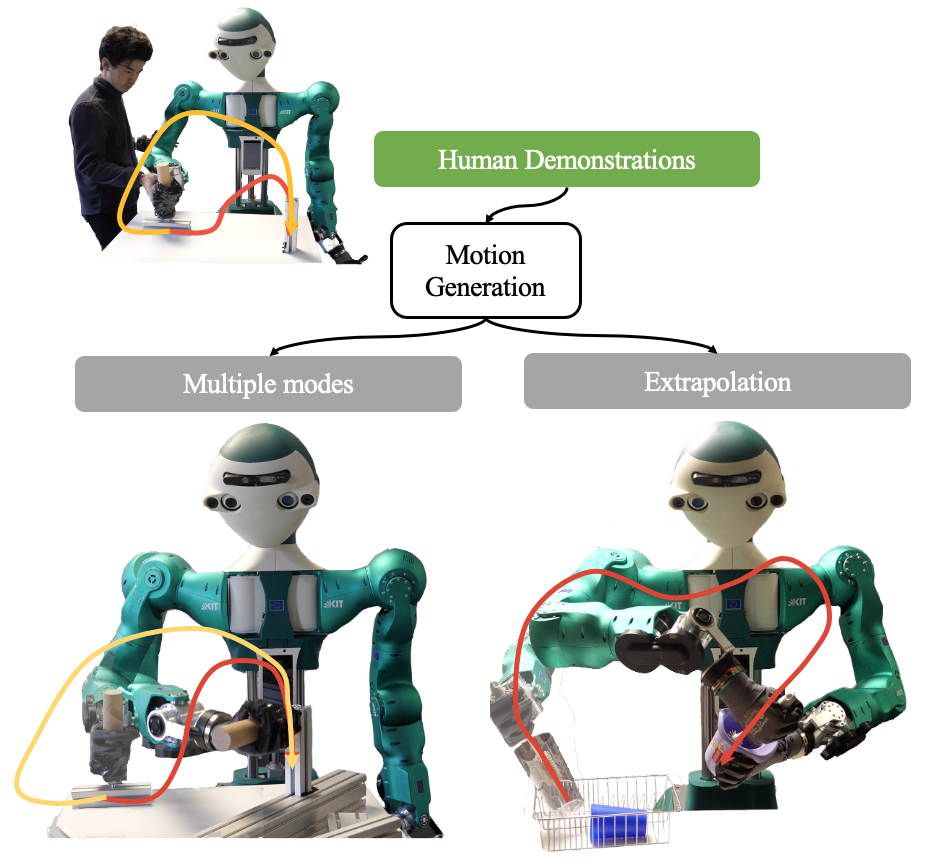} 
  \caption{Two challenges of a motion generation system: 1) multiple modes in human demonstrations; 2) extrapolation to task queries outside of the training range. }
 \label{fig:open}
 \vspace{-0.5cm}
\end{figure}

Several approaches in the literature have been proposed for learning a mapping from task queries to parameterized motion representations such as dynamic movement primitives (DMPs) (see~\cite{Ude2010, Forte2012, DaSilva12, Pahic2018} and \cite{Zhou2020}). Different regression models are applied for learning this mapping, such as locally weighted regression, Gaussian process regression, autoencoders, and mixture density network (MDN). The generalization capability of the motion generation model is dependent on the properties of regression methods. For example, compared to other methods, MDN can handle the multiple modes problem as shown in our previous work~\cite{Zhou2020}. However, like other methods, it cannot handle the extrapolation where new task queries are far from the training queries' range. 

To solve the multiple modes problem, instead of a GMM-based method such as MDN, we can convert a multiple mode distribution into a single mode distribution.  Here, one popular model is variational auto-encoder (\cite{Kingma2014}) (VAE), whose encoder maps data to a latent distribution close to a Gaussian distribution and the decoder maps the samples from latent distribution to the original space.  An alternative solution is the normalizing flow models, a set of invertible mappings that transform from arbitrary distributions to a simple distribution (see~\cite{Kobyzev2020} for a review).  These models are directly trained with likelihood costs and the change of variable formula.  Among these models,  the real valued non-volume perserving transformation (RealNVP) was proposed in~\cite{Dinh2016} and allows integrating arbitrary functions or neural networks and still guarantees invertibility with the unique structure.
VAEs and RealNVPs were extended separately to their conditional versions, such as in~\cite{Sohn2015, Ardizzone2019} to consider task queries. However, because of the poor regularization of neural networks, they can nearly extrapolate to queries outside the training queries' range. Hence, they cannot solve the problems mentioned above.

Task-parameterized Gaussian mixture model (TP-GMM), presented in~\cite{Calinon2013, Calinon2016} is a popular method to improve extrapolation ability. It restricts task queries to location and orientation of local frames and maps the task space trajectories to their representations in local frames. Then, Gaussian mixture models (GMMs) are learned for the distribution of these local representations. During the execution, TP-GMMs combine those GMMs and generates motions for new task queries.

However, the combination of Gaussian components from local frames distorts the generated global trajectories, especially when they are far away from the those in human demonstrations. In~\cite{Alizadeh2014, Huang2018, Sena2019}, the authors solve this problem by introducing a variance division factor to make the low variance component more valuable in Gaussian multiplication. The assumption is that the local frame with a lower variance is more important than the one with a higher variance. In~\cite{Sena2019}, the variance division factor $\alpha$ for each frame is dependent on the variance of their local trajectories $\mSig$.  For the $k$-th local frame, 
\begin{equation}\label{eq:var_weight}
	\alpha_k = \frac{\vert\vert \mSig_k^{-\gamma} \vert\vert}{\sum_{k=1}^K \vert\vert \mSig_k^{-\gamma} \vert\vert},
\end{equation}
where $\gamma$ is a parameter that determines how sensitive the value is to the variance.

The low variance assumption only works when task queries restrict the right motion trajectories, \ie the right demonstrations, in a small area of the task space. However, some task queries, such as an obstacle's location, do not require the motion trajectories to go through a small area. Hence, the proposed methods ignore such task queries. Moreover, TP-GMM based methods cannot solve multiple modes problem. 

\section{Our Approach}

Instead of learning GMMs for local frames as in TP-GMMs, we associate each local frame with a latent distribution given by a RealNVP. As discussed, RealNVPs can convert arbitrary, \ie multiple mode, distribution into a single mode distribution in the latent space. Compared to the approximate inference of VAEs, RealNVPs enable exact inference and sampling in the latent space with an invertible and stable mapping. This makes RealNVPs predestined to solve the multiple modes problem in motion generation

Like all other general purpose regression models, neural networks can hardly handle extrapolation. Hence, we avoid to provide task queries as inputs to the neural networks ever used in the model. Like TP-GMMs, we restrict the task queries to the location and orientation of local frames. However, unlike TP-GMMs that obtain a global motion trajectory with Gaussian multiplication, we consider that the robot should learn to shift its attention to different local frames during task execution. When attention is on one local frame, the motion trajectory is generated by the corresponding RealNVP. 
 
\begin{figure}[t]
  \centering
    \includegraphics[width=0.95\columnwidth]{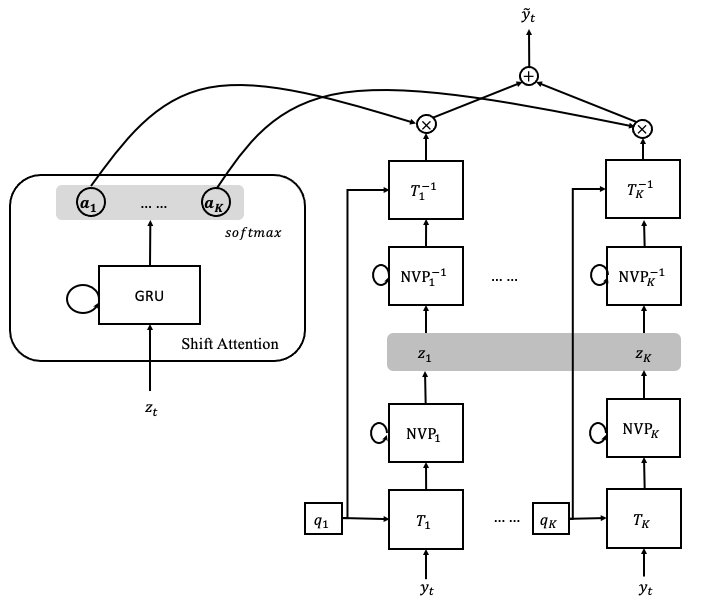} 
  \caption{SALaT: Shift Attention Latent Transformation. \textbf{Left:} a recurrent shift attention model based on GRUs outputs the attention weights $\{a_k\}_{k=1}^K$. \textbf{Right:} each local frame is associated with a linear transformation $T$ followed by a recurrent RealNVP.}
 \label{fig:salat}
\end{figure}

We propose a shift attention latent transformation (SALaT). As shown in~\autoref{fig:salat}, it consists of two parts: latent transformation and shift attention model. In the following, we first introduce latent transformation based on RealNVPs that transform trajectories to a latent distribution. Then, we introduce the attention model to combine these local models and explain how it is learned.

\subsection{Latent Transformations}

Before latent transformations, we first transform all demonstration trajectories into local frames. Each task query $\mq$ is associated with the transformations of $K$ local frames:
\begin{equation}
	T(\mq) = \left\lbrace T_k = (\mA^k, \mb^k)\right\rbrace_{k=1}^K,
\end{equation}
where $\mA^k$ and $\mb^k$ are the rotation and translation of the $k$-th local frame. From $N$ demonstrations, we get $N$ local trajectories for each local frame.

\begin{figure}[t]
	\centering
	\includegraphics[width=\columnwidth]{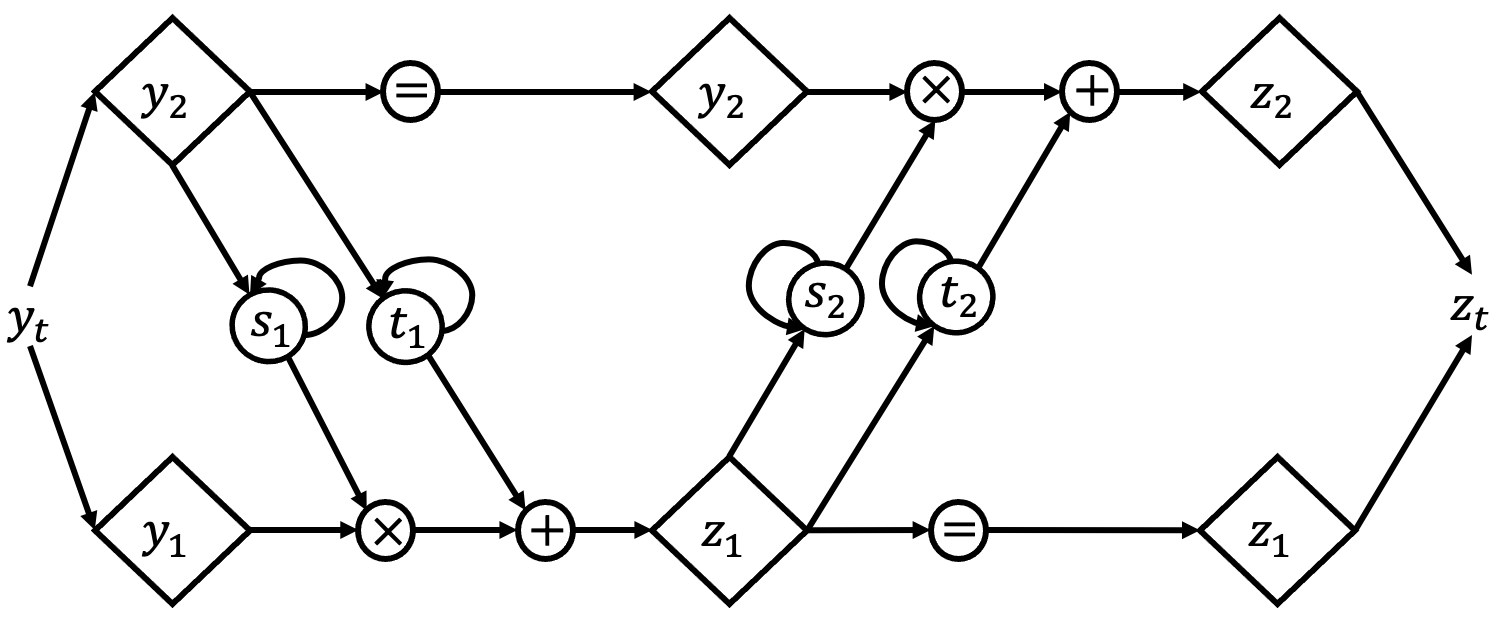} 
	\caption{The structure of a recurrent RealNVP. The scaling and translation networks $s_1, s_2, t_1, t_2$ are Bi-GRUs.}
	\label{fig:nvp}
	\vspace{-0.5cm}
\end{figure}

For the latent mapping, we split the local trajectory point at each timestamp into two parts such that $\my = [\my_1, \my_2]$. According to the RealNVPs shown in~\autoref{fig:nvp}, the latent transformation is 

\begin{equation}\label{eq:forward_nvp}
\begin{array}{ccc}
\mz_1 & = & \my_1 \odot \exp (s_1(\my_2))+ t_1(\my_2); \\
\mz_2 & = & \my_2 \odot \exp (s_2(\mz_1)) + t_2(\mz_1),
\end{array}
\end{equation}

where $s_1$, $s_2$ and $v_1$, $v_2$ are scaling and translation functions and $\odot$ is the Hadamard product or element-wise product. The inverse RealNVPs are calculated with
\begin{equation}\label{eq:backward_nvp}
\begin{array}{ccc}
\my_2 & = & (\mz_2 - t_2(\mz_1)) \odot \exp (-s_2(\mz_1)); \\
\my_1 & = & (\mz_1 - t_1(\my_2)) \odot \exp (-s_1(\my_2)).
\end{array}
\end{equation}
The advantage of a RealNVP is that its invertibility is independent of the structure of the scaling and translation functions. To consider the temporal state of the trajectory, we use bi-directional gated recurrent units (Bi-GRUs) (\cite{Schuster1997}, \cite{Cho2014}) to model these functions.

To train RealNVPs, we directly calculate the likelihood of the data with the change of variable formula, namely
\begin{equation}\label{eq:change_of_variable}
p(\my) = p_Z(f(\my)) \left\vert det \left(\frac{\partial f(\my)}{\partial \my^T} \right) \right\vert,
\end{equation}
where $f$ is the latent mapping shown in~\autoref{fig:nvp} and
\begin{equation}
det \left(\frac{\partial f(\my)}{\partial \my^T} \right) = \exp \left( \sum_{j_1} s_1(\my_2) + \sum_{j_2} s_2(\mz_1) \right).
\end{equation}
Hereby, $j_1$ and $j_2$ indicate the dimension of neural network outputs.  We can assume that the trajectory points at different timestamps are independent of each other, and use a negative log-likelihood cost function such that
\begin{equation}
L_{nvp} = - \sum_{i=1}^T \left(\log p_Z(f(\my)) + \sum_{j_1} s_1(\my_2, i) + \sum_{j_2} s_2(\mz_1, i) \right),
\end{equation}

where $T$ is the sequence length of the Bi-GRUs. The latent distribution $Z \sim \mathcal{N} (0, \mSig_0)$ follows a standard multi-dimensional Gaussian distribution.

This cost, however, does not consider the covariance between trajectory points at two different timestamps. Thus, it generates latent trajectories that might not be smooth in the latent space. To consider the covariance between timestamps, we construct a Gaussian distribution with a covariance matrix $K$ such that 
\begin{equation}
K_{i,j} = k(x_i,x_j) = \sigma_f^2 \exp \left( - \frac{(x_i - x_j)^2}{2 l^2} \right)
\end{equation}

where $x$ is a temporal factor and the hyper-parameters are predefined such that $\sigma_f=1$, $l=1$. The distribution $p_{\Xi}(\xi_{\mz}) \sim \mathcal{N}(0, K)$ is similar to a Gaussian process. Training RealNVPs with the cost function
\begin{equation}\label{eq:nvp_gp}
L_{nvp}= -\log p_{\Xi}(\xi_{\mz}) - \sum_{i=1}^T \left(\sum_{j_1} s_1(\my_2, i) + \sum_{j_2} s_2(\mz_1, i) \right)
\end{equation} 
gives better results than using the previous cost according to our experience.


\subsection{Recurrent Attention Model}

After training RealNVPs for local frames, we sample or use the mean latent trajectory from the standard latent distribution.  The attention model determines how much attention is paid on each local frame at each timestamp. As shown in~\autoref{fig:salat}, we use GRUs followed by a softmax layer to generate attention weights $\{a_k\}_{k=1}^K$ for $K$ local frames. With these attention weights, the generated trajectory point at a timestamp $t$ is

\begin{equation}
\hat{\my}_t = \sum_{k=1}^K a_{k,t}\cdot T_k^{-1} \left(\cdot NVP_k^{-1}(\mz_{k,t}) \right),
\end{equation} 
where $\{\mz_{k, t}\}_{k=1}^K$ are corresponding latent trajectory points.


The cost function for training the attention model consists of three different parts: \emph{variance weighted reproduction cost, attention distribution cost and smoothness cost}. The variance weighted reproduction cost guarantees that the generated trajectories by the model are similar to demonstrations especially in  low variance regions. However, it could ignore local frames containing high variance local trajectories. To solve this problem,  we introduce the attention distribution cost that distribute the attention according to the assumption that the robot should pay attention to one local frame at one time but needs to distribute attention equally to all local frames over time.  However, this cost introduces high acceleration in the generated trajectories. To guarantee the smooth transition, a smoothness cost is introduced.


\subsubsection{Variance Weighted Reproduction Cost}

In~\cite{Sena2019}, the author suggested a variance weighted reproduction cost to determine the parameter $\gamma$ in~\autoref{eq:var_weight}. We use a similar cost for the attention model. For $N$ demonstrations, we have $N$ local trajectories $\{\xi^n\}_{n=1}^N$ in each local frame. We calculate $K$ variance values for the trajectory points at timestamp $t$  and find the minimal value $v_t$ such that
\begin{equation}
v_t = \min_{k \in K} Var\left(\{ \xi_{k,t}^{n} \}_{n=1}^N \right).
\end{equation}
The variance weight is calcuated by
\begin{equation}
w_t = \frac{1}{v_t + \epsilon},
\end{equation}
where $\epsilon=0.01$ is used in all our experiments. The reproduction cost is
\begin{equation}
L_{reprod}= \sum_{n=1}^N \sum_{t=1}^T w_t \vert\vert \xi_t^n - \hat{\xi}_t^n \vert\vert^2
\end{equation}
with $\xi$ and $\hat{\xi}$, the demonstrated and generated trajectories.

\subsubsection{Attention Distribution Cost}

The variance weighted reproduction cost penalizes the reproduction errors in low variance regions but ignores totally the local frame containing high-variance local trajectories such as an obstacle mentioned before. To solve this problem, we assume that the attention should be roughly equally distributed to all $K$ local frames during the task execution and end up with an attention distribution cost such that
\begin{multline}
	L_{traj}  =  - \log \frac{1}{K}  +\\
\frac{1}{N} \sum_{n=1}^N \sum_{k=1}^K \left(\frac{\sum_{t=1}^T a_{k,t}^n}{T} \right) \cdot \log \left(\frac{\sum_{t=1}^T a_{k,t}^n}{T} \right)
\end{multline}
where $a_{k,t}^n$ is the attention on $k$-th local frame at the timestamp $t$ in $n$-th demonstration, and $(\sum_{t=1}^T a_{k,t}^n ) / T$ is the average attention on $k$-th local frame in $n$-th demonstration.

Minimizing this cost function distributes the attention equally to different local frames over time. However, it can also result in an equal distribution of attention at each timestamp, which weakens the model's extrapolation ability.  Considering the case when attention is distributed equally to different local frames at each timestamp, the model cannot extrapolate when local frames are far away from each other because the unseen relative position of local frames unpredictably distorts the trajectory shape. Hence, we introduce another cost such that
\begin{equation}
L_{point} = \frac{1}{N \cdot T} \sum_{n=1}^N \sum_{t=1}^T \sum_{k=1}^K \left(-a_{k,t}^n \log a_{k,t}^n\right).
\end{equation} 
Minimizing this cost distributes the attention to only one local frame at each timestamp. The attention distribution cost is the summation of both costs that
\begin{equation}
L_{dist} = w_{traj} L_{traj} + w_{point} L_{point}.
\end{equation}
In our experiments, $w_{traj}=10$ and $w_{point}=1$.

The attention distribution cost corresponds to the assumption that a robot should pay attention to one local frame at each timestamp but distribute its attention to all local frames equally during the whole task. 

\subsubsection{Smoothness Cost}
However, the attention distribution cost results in a rapid shift of attention from one local frame to another and, thus, a large acceleration in generated trajectories. To solve this problem, we introduce the cost to penalize the rapid changes in the trajectory such that
\begin{equation}
L_{smooth} = \frac{1}{N} \sum_{n=1}^N\sum_{t=1}^{T-1} \vert\vert  \hat{\my}_{ t+1, n} - \hat{\my}_{t,n} \vert\vert^2.
\end{equation}
The smoothness cost guarantees the smooth transition from one local model to another. 
The total cost for training the attention model is the sum of all three costs such that
\begin{equation}
L_{att} = L_{reprod} + L_{dist} + L_{smooth}.
\end{equation}

In many applications, local trajectories already follow Gaussian distribution such as in the docker experiment (see~\autoref{subsec:docking}).  In this case, we can skip the latent transformation and directly use the shift attention model. We call such simplified version of SALaT the shift attention linear transformation (SALiT). However, as shown in other tasks than the docker experiments, a Gaussian is not enough to represent the local trajectory distribution, where SALiT fails to generate valid trajectories.


\subsection{Task Execution with SALaT}

For task execution, we can either draw latent trajectories from the latent distribution or use its mean trajectory and transform them to local trajectories according to new task queries. The attention model is applied to combine these local models and generate global trajectories to accomplish the task.

Sampling from the latent distribution does not guarantee the generated trajectories' soundness because the resulting samples can still be away from the mean. To evaluate the success rate in simulated experiments, we instead selected the mean latent trajectory. In~\autoref{subsec:rob_exp}, we use samples to show that SALaT can generate trajectories associated with different modes.

\section{Evaluations} \label{sec:evaluation}
\begin{figure}[t]
	\centering
	\includegraphics[width=0.9\columnwidth]{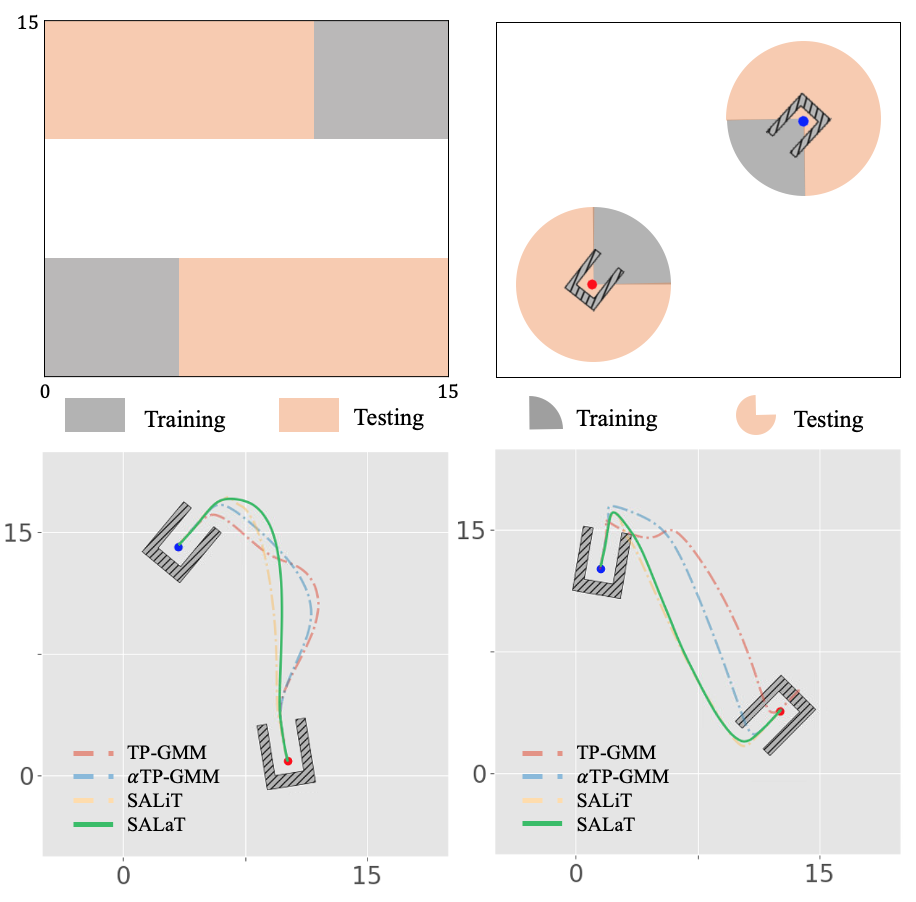} 
	\caption{Dockers Experiment. \textbf{Top:} the training and testing data for the experiment. \textbf{Bottom:} two examples for extrapolation.}
	\label{fig:docking}
	\vspace{-0.5cm}
\end{figure} 

To evaluate our method, we conduct simulated and real experiments. Inspired by~\cite{Calinon2013, Sena2019}, we construct a docker experiment, where a valid trajectory connects the start and goal dockers without collision. It is an abstraction of many robot applications as shown in~\autoref{subsec:rob_exp}.  To introduce multiple modes and increase the task complicity, we add one obstacle and construct a docker-obstacle experiment, where a valid trajectory should also go around the obstacle. To make the task even more difficult, we change the goal docker to a tunnel and construct a docker-obstacle-tunnel experiment, where a valid trajectory should go through the tunnel and come back to the start docker. We compare SALaT with previous approaches based on TP-GMMs (\cite{Calinon2016} and~\cite{Sena2019}) in all three experiments. 

In robot experiments, we construct a task where the robot should slide a tool out of a aluminium profile and insert it into another one. Since it allows two different kinds of sliding motions, it rasises multiple modes problem. After being trained on a limited number of demonstrations, the SALaT can extrapolate to new task queries and generate motions from different modes.


\subsection{Docker Experiment}\label{subsec:docking}
\begin{figure}[t]
	\centering
	\includegraphics[width=0.9\columnwidth]{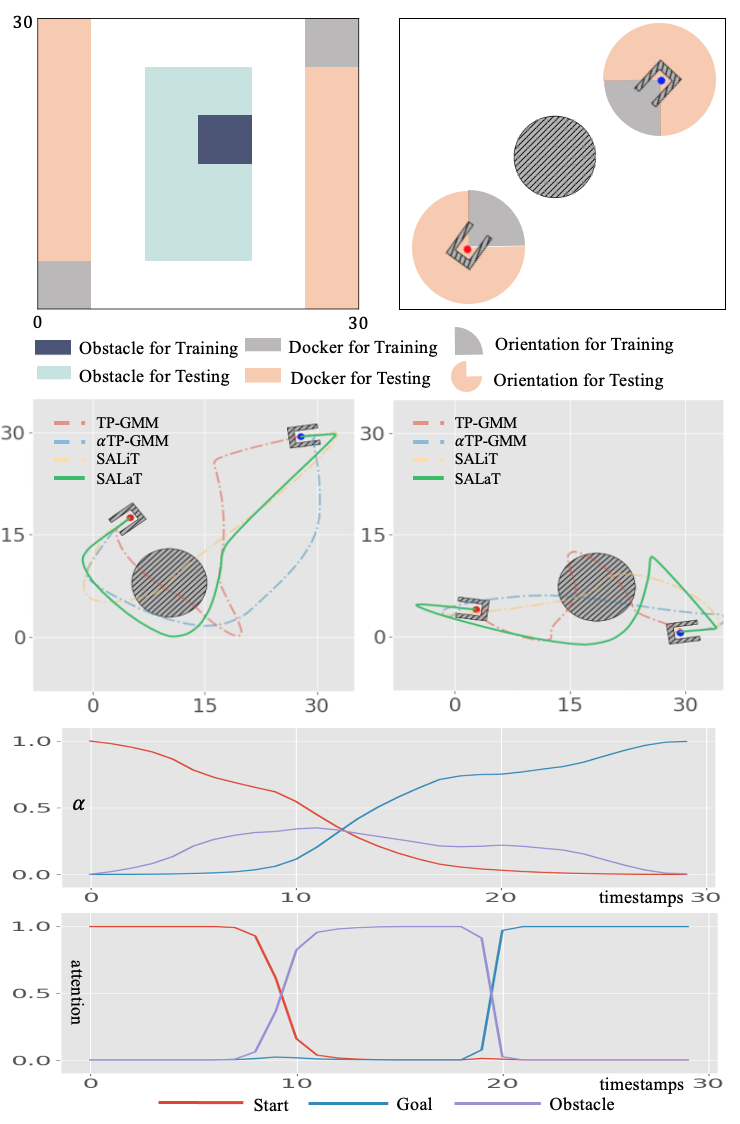} 
	\caption{Docker-Obstacle Experiment.  \textbf{First row:} training and testing dataset for the docker-and-obstacle experiment. \textbf{Second row:} two testing examples for extrapolation. \textbf{Third row:} the change of $\alpha$ in $\alpha$TP-GMM. \textbf{Last row:} the outputs of the shift attention model.}
	\label{fig:dockingObs}
\end{figure}

\begin{table}
\centering
\begin{tabular}{ |p{1cm}|c|c|c|c| } 
 \hline
  Exp & TP-GMM(\cite{Calinon2013}) & $\alpha$TP-GMM(\cite{Sena2019}) & SALiT & \textbf{SALaT} \\ 
  \hline
 1 & 0.42 & \textbf{0.60} & 0.55 & \textbf{0.60} \\ 
  \hline
 2  & 0.06 & 0.25 & 0.0 & \textbf{0.52} \\ 
 \hline
3 & 0.10 & 0.09 & 0.0 & \textbf{0.40} \\ 
\hline
\end{tabular}
\caption{Success rate of methods for different experiments. Experiments are \textbf{1:} docker experiment, \textbf{2:} docker-obstacle experiment, \textbf{3:} docker-obstacle-tunnel experiment}
\label{tab:success_rate}
\vspace{-0.5cm}
\end{table}
For the docker experiment, we collected $20$ training data on a tablet and tested on $100$ new task queries. As shown in the top of~\autoref{fig:docking}, 
the testing queries are uniformly sampled from the locations and the orientations that do not appear in the training dataset. Two examples are shown in the bottom of~\autoref{fig:docking}. The success rate is calculated by counting the number of successful trials in $100$ tests. As shown in the first row of~\autoref{tab:success_rate}, all three methods $\alpha$TP-GMM, SALiT and SALaT performed equally well in the task. As mentioned before, since demonstrations in the docker experiment already follow a Gaussian distribution, a latent representation does not improve the performance a lot.  Furthermore, it meets the assumption made by $\alpha$TP-GMM that the local frame containing low variance trajectories is important at each timestamp. For the attention model, it just shifts the attention simply from the start docker to the end docker.

\subsection{Docker-Obstacle Experiment}\label{subsec:obs_docking}

\begin{figure}[t]
	\centering
	\includegraphics[width=0.95\columnwidth]{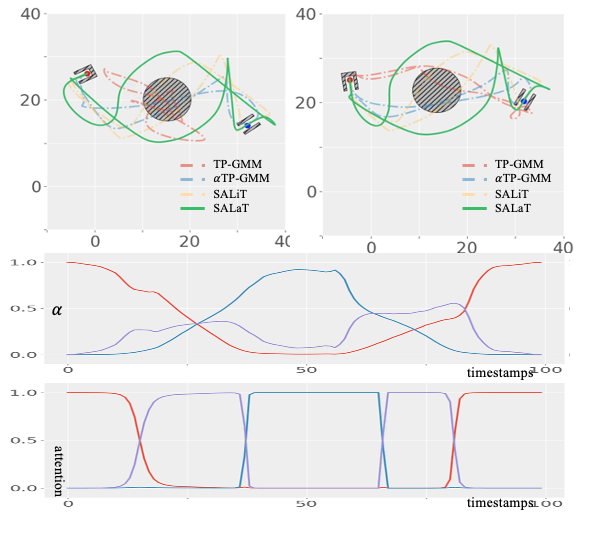} 
	\caption{Docker-Obstacle-Tunnel Experiment. \textbf{First row:}  testing examples for extrapolation. \textbf{Second row:} the change of $\alpha$TP-GMM. \textbf{Third row:}the outputs of the shift attention model. }
	\label{fig:dockingReturn}
	\vspace{-0.5cm}
\end{figure}
\begin{figure*}[t]
	\centering
	\includegraphics[width=0.9\textwidth]{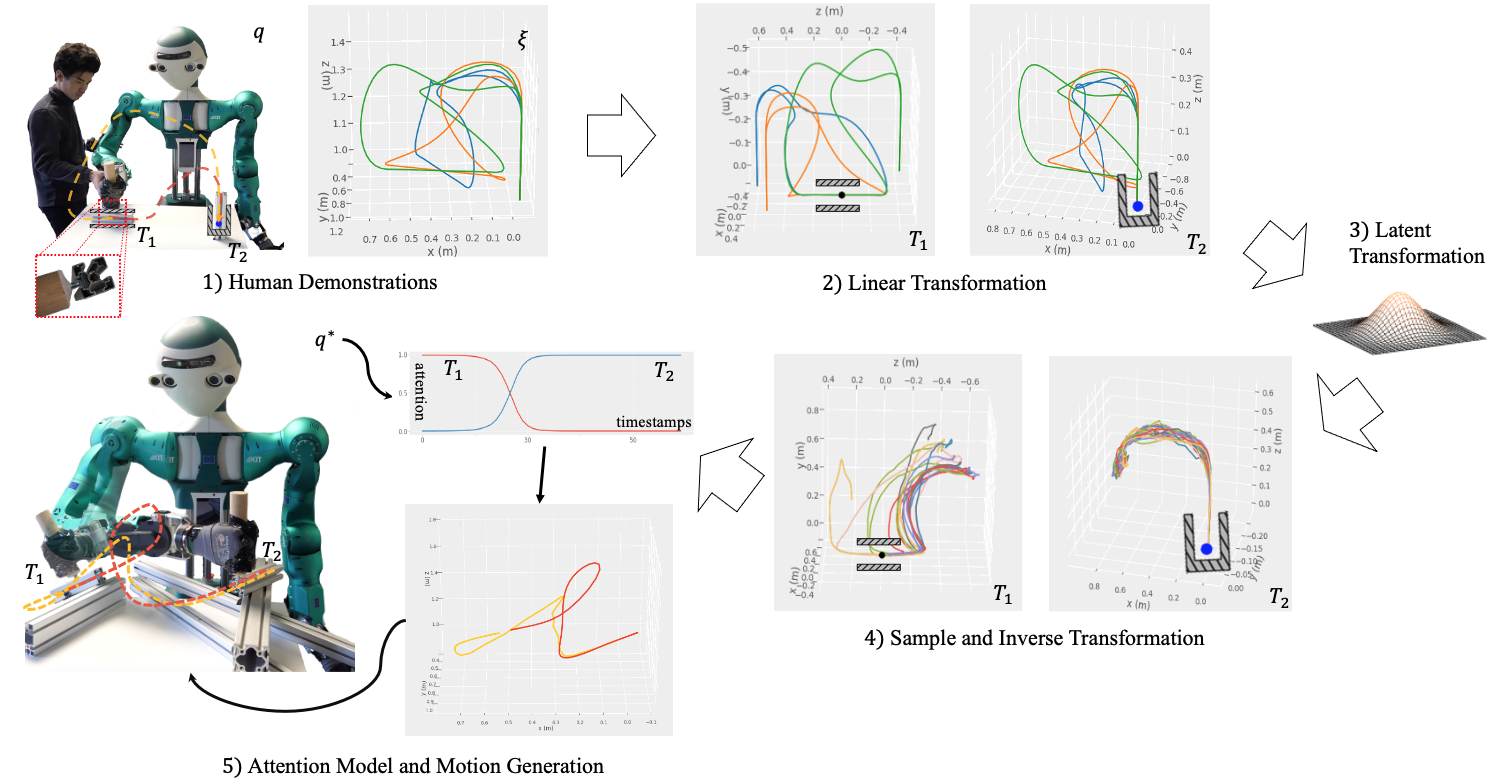} 
	\caption{Robot Experiments with SALaT.  }
	\label{fig:rob_exp}
	\vspace{-0.5cm}
\end{figure*}

We add one obstacle in between two dockers to introduce multiple modes in the task. A human can draw the curves that go around the obstacle from either its left or right side. We collected $30$ data for training and used $100$ testing queries. As shown at the top of~\autoref{fig:dockingObs}, the extrapolation test dataset has no overlap with the training dataset.  In the two examples shown in the second row of~\autoref{fig:dockingObs}, all methods except SALaT fail to generate a collision-free trajectory because they average both modes. As shown in~\autoref{tab:success_rate}, SALaT performs better than other methods.  

The third row of~\autoref{fig:dockingObs} describes the change of $\alpha$ given by~\autoref{eq:var_weight}, the $\alpha$TP-GMM ignores the obstacle all the time (see the blue curve), which is another reason why it does not work well in this experiment. It can still have a $25\%$ success rate without considering the obstacle because it is randomly placed and might not necessarily be between two dockers. In contrast, both SALaT and SALiT pay attention to the obstacle even though the trajectories' variance is high. Since the SALiT uses the average of the trajectories that go through the obstacle, it has a zero success rate.

\subsection{Docker-Obstacle-Tunnel Experiment}
To further evaluate the model, we introduce a tunnel into the experiment. A successful trial means that the trajectory goes through the tunnel and returns to the start docker while avoiding collisions with all obstacles (see the top diagrams in~\autoref{fig:dockingReturn} for two testing examples).  The training and testing queries have similar ranges, as in the previous experiment (see the top plots in~\autoref{fig:dockingObs}). We collected $30$ demonstrations and randomly sampled $100$ testing queries. The attention should be paid twice to the obstacle. As shown with the purple curves at the bottom of~\autoref{fig:dockingReturn},  a SALaT model successfully learns how to shift attention to realize the successful task execution. The $\alpha$TP-GMM also changes the parameter $\alpha$ accordingly because human demonstrations accidentally have relatively low variance when getting near to the obstacle. However, $\alpha$ is still not significant enough to generate correct trajectories around the obstacle. As shown in~\autoref{tab:success_rate}, the SALaT model outperforms others. 

\subsection{Robot Experiment}\label{subsec:rob_exp}

We conducted the robot experiment on the humanoid \armarVI (\cite{Asfour2018b, Asfour2019}). We use the aluminum profiles and implement an experiment similar to the docking problem, where the robot should slide a tool out of one profile and insert it into another one. The tool can be taken out of the profile by sliding it from either side but cannot be pulled directly out, as shown in the most top-left picture of~\autoref{fig:rob_exp}, which requires multiple modes. To evaluate extrapolation, we fixed the goal profile pose and only rotated the initial profile on the table to collect human demonstrations, and sampled the testing queries from the whole working space where the robot's arm is reachable. To evaluate multiple modes, we intentionally demonstrated two different motions for one query.

The~\autoref{fig:rob_exp} shows a complete process of using SALaT for real robot tasks.   1) We collect $16$ demonstrations $\xi$ for $8$ task queries $\mq$s. Only $6$ trajectories for $3$ queries are shown in the plots for clarity. 2) All motion trajectories are transformed into two local frames $T_1$ and $T_2$. 3) RealNVPs are trained on local trajectories for each local frame; 4) We draw latent trajectories from the latent distribution and transform them back to local frames. 5) We learn the attention model and generate motion trajectories for any new task queries $\mq^*$ and finally execut trajectories on the robot.  With good extrapolation ability, the learned SALaT for the aluminum profile task can be directly used for other tasks such as inserting a cup brush into a container as shown in~\autoref{fig:open}.




\section{Conclusion}

In this paper, we introduce a new model,  the shift attention latent transformation (SALaT), that consists of local latent transformations and an attention model. The local latent transformations solve the multiple modes problem that exists in human demonstrations, while the attention model improves the extrapolation ability for motion generation. We can consider the local latent representation given by RealNVPs as the vocabulary, based on which the attention model generates  motion trajectories for the task. 
However, like TP-GMM, SALaT can only work in the robot task space and requires further transformations for the mapping between different spaces. In the future, we will solve how to automatically learn the transformations to replace the linear transformations in SALaT. We also want to explore the possibility of the model in the context of reinforcement learning.

\addtolength{\textheight}{-12cm}   

\bibliographystyle{IEEEtran} 
\bibliography{references}

\end{document}